\documentclass[journal,letterpaper]{IEEEtran}

\usepackage{cite}
\usepackage{amsmath,amssymb,amsfonts}
\usepackage{algorithmic}
\usepackage{graphicx}
\usepackage{textcomp}
\usepackage{xcolor}
\usepackage{booktabs} 
\usepackage{url}

\title{Rank-Aware Quantile Activation for Motion-Robust Crop Segmentation in UAV Imagery}

\author{
    \IEEEauthorblockN{Abinav Kiran, Sravan Danda, Aditya Challa, Sougata Sen, and B S Daya Sagar, \textit{Senior Member, IEEE}}
}

\begin{document}

\maketitle

\begin{abstract}
Motion blur from high-speed UAV acquisition degrades semantic segmentation on rare texture-dependent classes with high agronomic value. Standard CNNs rely on high-frequency magnitude features that blur destroys, causing statistical erasure of minority signals. We propose Dual Quantile Activation (QAct), a rank-aware block replacing magnitude gating with instance-level rank normalization. Evaluated on Agriculture-Vision 2021 across zero-shot and blur-supervised regimes at multiple severities, QAct is the dominant architectural factor: it delivers consistent mIoU gains over ReLU across both regimes and all severities, with strongest gains on rare structural and texture-dependent classes. Some dominant classes (water, planter skip) show mixed per-class performance under distillation. At moderate blur, zero-shot QAct outperforms distillation-trained ReLU; across all severities, Distill-QAct achieves best performance, confirming rank-aware activation and blur-domain training are complementary robustness sources.
\end{abstract}

\begin{IEEEkeywords}
Precision Agriculture, Semantic Segmentation, Unmanned Aerial Vehicles (UAV), Motion Blur, Quantile Activation, Deep Learning.
\end{IEEEkeywords}

\section{Introduction}
\IEEEPARstart{S}{caling} precision agriculture demands rapid, cost-effective field monitoring. Unmanned Aerial Vehicles (UAVs) have emerged as the primary platform for crop scouting, weed mapping, and yield estimation at field scale~\cite{chiu2020agriculture}: they offer high temporal revisit rates and sub-meter ground sampling distance that satellites cannot match economically. However, the operational speed required for coverage efficiency introduces a physical constraint that existing segmentation benchmarks have largely ignored: severe 6-DOF motion blur. At realistic scouting speeds, the camera's exposure integral smears fine-grained agronomic anomalies—localized weed clusters, endrow compaction zones, nutrient-deficiency patches—into elongated, low-contrast streaks. These are precisely the texture-dependent, rare classes with the highest agronomic value. Standard Convolutional Neural Networks (CNNs), trained on pristine satellite orthoimagery, rely on high-frequency magnitude cues that motion blur physically destroys, causing catastrophic segmentation collapse on the classes that matter most.

Current state-of-the-art segmentation models (e.g., HRNet~\cite{sun2019deep}, DeepLabV3+~\cite{chen2018encoder}) operate on the assumption of availability of high-quality input data. When deployed on blurred UAV imagery, their performance collapses, particularly for small objects or texture-dependent classes. Furthermore, traditional blind deblurring~\cite{kupyn2018deblurgan, kupyn2019deblurganv2, xiang2024deep} is computationally prohibitive for real-time onboard inference and often fails due to the spatially varying nature of 6-DOF (Degree of Freedom) camera motion.

\begin{figure*}[t]
    \centering
    \includegraphics[width=\textwidth]{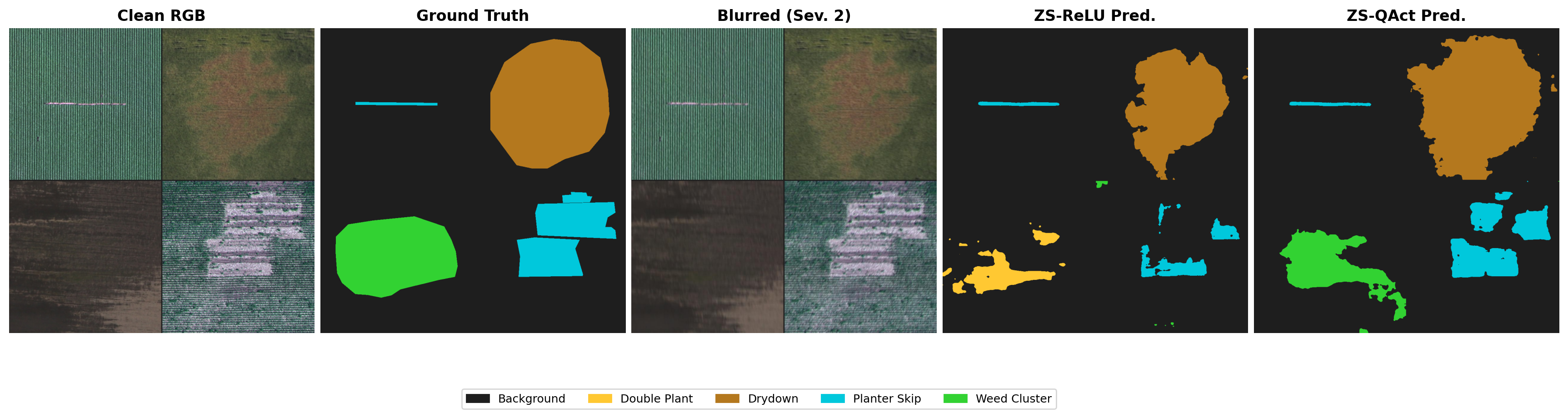}
    \caption{\textbf{Zero-Shot Architectural Robustness Under Moderate Blur (Severity 2).}
    Each panel is a $2{\times}2$ mosaic of four independent Agriculture-Vision~2021 validation tiles, chosen so that each tile is dominated by a single agronomic class with background-only borders to avoid seam artifacts: top-left \textit{background}, top-right \textit{drydown} (majority), bottom-left \textit{weed\_cluster} (minority, texture-dependent), bottom-right \textit{planter\_skip} (minority, thin linear). Each tile is inferred separately at Severity~2 blur; predictions are stitched into the same layout. Both models were trained on \emph{clean} data only — no blur exposure during training. ZS-ReLU collapses on the minority tiles: \textit{weed\_cluster} pixels fall into the ReLU dead-zone and are erased, and the thin \textit{planter\_skip} rows disappear almost entirely. ZS-QAct, identical in architecture and training data apart from the Dual QAct block, recovers both rare classes through rank-aware variance renormalization while preserving the majority \textit{drydown} prediction. This isolates the contribution of the architectural inductive bias from any blur-supervised training: rank-awareness alone is sufficient to survive operationally realistic blur on exactly the rare, spatially sparse classes most vulnerable to statistical erasure.}
    \label{fig:typology_ablation}
    \end{figure*}
    
In this work, we propose a shift from purely magnitude-based feature learning to \textit{rank-aware} feature learning for robust agricultural segmentation. The core of our approach is the \textit{Dual Quantile Activation (QAct)} block~\cite{challa2025quantile}: a residual activation scheme that replaces magnitude gating with instance-level rank normalization, structurally renormalizing the spatial variance of blurred features so that minority class signals survive the activation layer intact. Crucially, QAct requires \textit{no blur exposure during training} — the architectural inductive bias alone provides the robustness. Detailed mathematical formulations are provided in Section~\ref{sec:main}.

To isolate the contribution of QAct from the contribution of blur-domain training data, we compare four variants spanning two orthogonal axes: activation type (ReLU vs.\ QAct) and training regime (zero-shot clean-only vs.\ blur-supervised EMA distillation). Our central empirical finding is that \textit{QAct is the dominant factor across both axes}: within the zero-shot regime, QAct gains range from +37\% to +66\% relative mIoU as severity increases; within the distillation regime, QAct provides an additional +8\% to +20\% relative gain on top of blur supervision. At moderate blur (Severity 2), zero-shot QAct even outperforms distillation-trained ReLU, confirming that the rank-aware inductive bias can substitute for blur-domain training data under operationally realistic conditions.

Our specific contributions are:
\begin{enumerate}
\item We adapt the \textit{Quantile Activation (QAct) layer}~\cite{challa2025quantile} for dense prediction tasks. By applying per-instance normalization~\cite{ulyanov2016instance} prior to rank mapping, we engineer a Residual Quantile Block that preserves the spatial structural hierarchy of agricultural features independently of global batch statistics.
\item We establish that QAct is the \textit{dominant architectural factor} for blur robustness, providing consistent gains in both zero-shot and distillation-trained settings across all blur severities, with especially large and reliable margins on rare structural and texture-dependent classes. The QAct gain within the zero-shot regime (+37\%--+66\% relative mIoU) demonstrates that the architectural inductive bias alone is sufficient under moderate blur; the additional QAct gain within distillation (+8\%--+20\%) confirms that QAct is complementary to, not redundant with, blur-domain training data.
\item We translate this architectural advance into a deployment-ready capability for UAV-based precision agriculture. Evaluated on the multi-modal Agriculture-Vision 2021 dataset~\cite{chiu2020agriculture} under a physically grounded 6-DOF blur simulation calibrated to operational DJI Agras flight envelopes, zero-shot QAct provides robust segmentation \textit{without requiring blur-augmented training data} — addressing a critical constraint in field-scale agricultural surveys, where paired clean/blurred acquisitions are impractical to collect at scale. The largest gains concentrate on rare structural (\textit{endrow}) and texture-dependent (\textit{weed cluster}) classes that drive variable-rate spraying and yield-loss assessment.
\end{enumerate}

\section{Evaluation Framework: Drone Physics}

\subsection{The Challenge of 6-DOF Motion Blur}
To rigorously evaluate the robustness of our proposed architecture, we must first model the realistic degradation conditions of UAV acquisition.  Unlike linear motion blur, often synthesized in computer vision (e.g., via simple kernel translation), drone imaging involves complex ego-motion. 
A UAV undergoes translation $(t_x, t_y, t_z)$ and rotation (roll, pitch, yaw) simultaneously during the camera's exposure time $T$.

Severe motion blur is simulated using a physically-realistic 6-DOF drone dynamics model at Severity 3 (representing realistic operational UAV speeds), introducing spatially variant directional smearing and photometric attenuation across both RGB and NIR channels.

Because the depth of the scene varies (even slightly in agriculture) and the camera rotates, the blur kernel is spatially variant. A weed at the edge of the frame suffers different smear trajectories than a crop row in the center. Consequently, no single deconvolution function can recover the sharp image globally. Additionally, each image is typically captured during a different drone flight. Thus, the test-time image distribution is effectively a complex mixture of diverse motion blur distributions, adding another layer of difficulty that breaks standard magnitude-dependent feature extractors.

\subsection{Simulation Setup for Validation}
We explicitly note that we do not propose this 6-DOF simulation as a novel contribution. Rather, we utilize it as a rigorous, physically grounded testbed to empirically validate our claims regarding rank-aware feature robustness.

To generate our evaluation data, we employ a simulation pipeline that maps blur severity levels (1 to 5) to physical flight parameters. For a given severity, we sample random angular velocities and translation vectors consistent with real-world drone dynamics. We generate the blurred image by accumulating $N$ sub-frames, warping the sharp ground truth according to the sampled trajectory.

Crucially, agricultural sensors capture multi-spectral data. We apply the \textit{same} temporal integration trajectory to both RGB and Near-Infrared (NIR) channels simultaneously. This preserves the geometric alignment of the blur across spectra, which is a critical requirement for multi-modal fusion networks.

\section{The Residual Quantile Block: Superimposing Rank onto Magnitude}
\label{sec:main}

We deploy a residual parallel activation scheme within our HRNet backbone, which we term the \textit{Dual QAct block}: a two-pathway structure combining a standard ReLU pathway (magnitude-preserving) with a $\text{QAct}_{2D}$ pathway (rank-preserving), summed with a learnable scale:
\begin{equation}
    F_{out}(X) = \text{ReLU}(X) + \lambda \cdot \text{QAct}_{2D}(X)
\end{equation}
where $\lambda$ is a learnable scaling parameter initialized to $0.1$. The $\text{QAct}_{2D}$ target is a standard normal $\mathcal{N}(0,1)$, superimposing a dense, zero-mean rank-signal onto the sparse rectified magnitude signal. For uncorrupted features ($X > 0$), the ReLU pathway dominates; when blur attenuates features into the dead-zone ($X \le 0$), the $\text{QAct}_{2D}$ pathway circumvents statistical erasure by populating it with variance-recovered rank data, ensuring gradients continue to flow for minority class signals.

\subsection{Statistical Erasure vs Distribution Alignment}
\begin{figure}[h]
    \centering
    \includegraphics[width=0.85\linewidth]{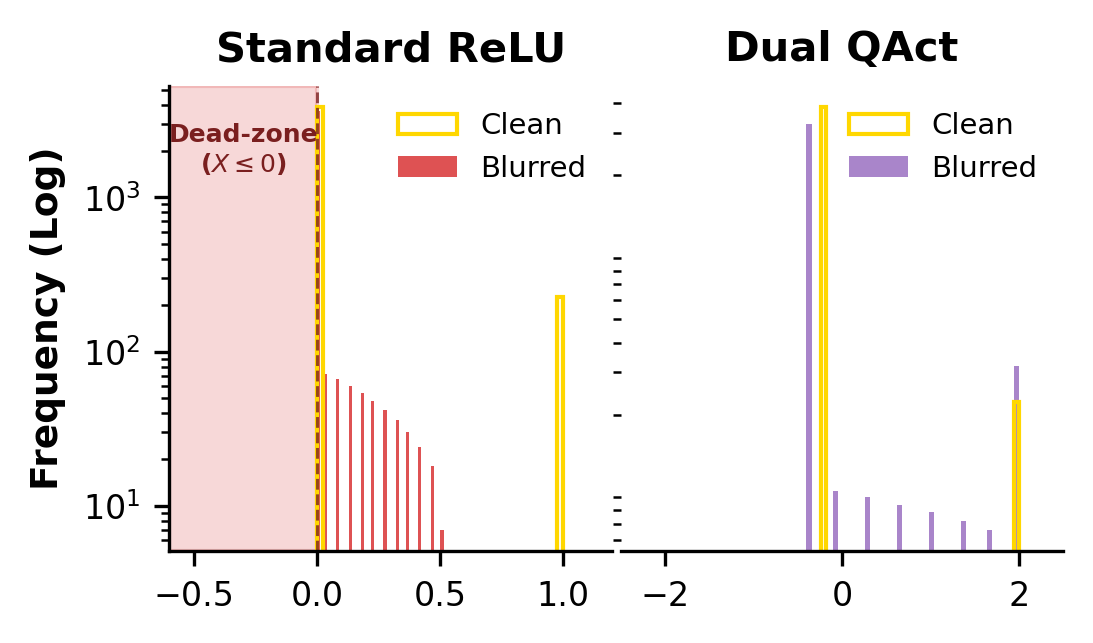}
    \caption{Feature activation densities (log scale) under clean (hollow yellow) and blurred (colored) conditions. \textbf{(Left)} ReLU: blur drags activations into the dead-zone ($x \leq 0$), erasing the minority class signal. \textbf{(Right)} QAct: per-instance renormalization preserves alignment with the clean distribution, maintaining viable gradient flow.}
    \label{fig:histogram_proof}
\end{figure}

Fig.~\ref{fig:histogram_proof} empirically confirms this mechanism: under severe blur, ReLU activations collapse into the dead-zone ($X \leq 0$), erasing the minority class signal entirely, while QAct's per-instance renormalization keeps the degraded distribution aligned with the clean signal boundaries, preserving gradient flow.

\subsection{Comparison: Blur-Supervised Distillation Baselines}
\label{subsec:distillation}

To contextualize QAct's zero-shot robustness, we compare against distillation-trained variants that enjoy privileged access to blurred imagery during training. Both distillation baselines employ a mean-teacher framework~\cite{tarvainen2017mean}: a teacher network (exponential moving average of the student weights) processes clean images, while the student is simultaneously trained on blur-augmented inputs. The student objective combines a task loss on both clean and blurred inputs with an $\mathcal{L}_1$ soft-target alignment loss~\cite{hinton2015distilling} between student(blurred) and teacher(clean) probability distributions, encouraging the student to align its blurred-input predictions to clean-input targets. We evaluate a standard ReLU backbone under this regime (\textbf{Distill-ReLU}) and a QAct backbone under the same setup (\textbf{Distill-QAct}).

This comparison isolates two orthogonal factors: activation type (ReLU vs.\ QAct) and training regime (zero-shot vs.\ distillation). If blur robustness were purely a data problem, distillation alone should close the gap between ReLU and QAct; if it were purely architectural, QAct should deliver the same relative gain regardless of training regime. Our results show both factors contribute, with QAct as the dominant and more consistent driver across severities and class typologies.

\section{Experiments and Results}

\subsection{Dataset and Evaluation Protocol}
\label{subsec:dataset}
We evaluate on the Agriculture-Vision 2021 dataset~\cite{chiu2020agriculture, chiu2020challenge}, using 8 valid semantic classes after discarding the ambiguous \textit{storm\_damage} class. The dataset exhibits severe class imbalance representative of real-world agricultural deployment: macro classes (\textit{drydown}: 12.33\%, \textit{nutrient\_deficiency}: 7.43\%) dominate the foreground pixel distribution, while rare structural and texture-dependent anomalies are exceedingly scarce (\textit{endrow}: 0.85\%, \textit{double\_plant}: 0.79\%, \textit{planter\_skip}: 0.19\%). These rare classes carry the highest agronomic value and are the most vulnerable to ReLU statistical erasure under blur.

We generate blurred evaluation sets at Severity 2, 3, and 4 using our 6-DOF simulation, spanning moderate to severe UAV motion. Severity 3 corresponds to realistic operational drone speeds (e.g., DJI Agras series at 5--10~m/s, 1/500s exposure). We evaluate four variants: \textbf{ZS-ReLU} (standard HRNet, clean training only), \textbf{ZS-QAct} (Dual QAct, clean training only — proposed), \textbf{Distill-ReLU}, and \textbf{Distill-QAct} (blur-supervised EMA distillation as described in Section~\ref{subsec:distillation}). All use HRNet-W18-Small as the backbone. Confidence intervals are 95\% bootstrap CIs (10{,}000 iterations) over per-image IoU values across the 18{,}334-image validation set, ensuring statistically grounded comparisons rather than point estimates.

On the unblurred validation set, the four models achieve clean-data mIoU of 0.356 (ZS-ReLU), 0.418 (ZS-QAct), 0.359 (Distill-ReLU), and 0.380 (Distill-QAct), confirming that neither QAct nor distillation training regresses clean-image performance. The published ResNet-101 baseline on Agriculture-Vision 2021~\cite{chiu2020challenge} reports 0.43 mIoU; HRNet-W18-Small is a lighter backbone and its clean-data performance falls in the expected range, establishing that the severe mIoU drops under blur are degradation-induced rather than a consequence of a weak baseline. As a ViT baseline, SegFormer-B1 achieves a clean-data mIoU of 0.406, comparable to HRNet-W18-Small; however, under blur it collapses to 0.351 at Severity 2 (-13.4\%) and 0.139 at Severity 4 (-65.8\%), demonstrating that self-attention alone does not confer robustness to motion blur and confirming that explicit architectural treatment is necessary.

\subsection{Per-Class Results Across Blur Severities}

\begin{figure*}[t]
    \centering
    \includegraphics[width=\textwidth]{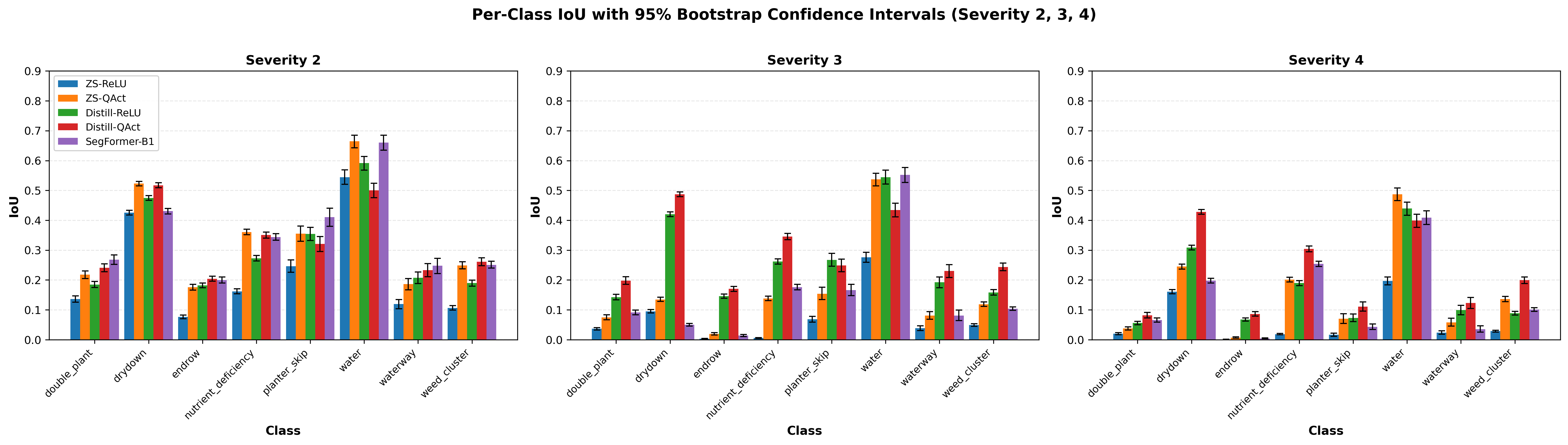}
    \caption{\textbf{Per-Class IoU with 95\% Bootstrap Confidence Intervals (Severity 2, 3, 4).}
    Each panel shows all 8 foreground classes with five grouped bars
    (ZS-ReLU, ZS-QAct, Distill-ReLU, Distill-QAct, and SegFormer-B1) and error bars representing
    95\% bootstrap CIs over 18{,}334 validation images.
    QAct provides broad gains over ReLU across both training regimes and all severities,
    with the largest and most consistent margins on rare structural and texture-dependent classes
    (\textit{endrow}, \textit{weed\_cluster}) most susceptible to ReLU statistical erasure.
    Some dominant classes (\textit{water}, \textit{planter\_skip}) show mixed ordering
    under distillation at moderate severities, reflecting that data-rich training
    can partially compensate for magnitude-based limitations on high-contrast classes.
    SegFormer-B1, a Vision Transformer baseline, shows comparable clean-data performance to HRNet but
    similarly degrades under blur, confirming that attention-based architectures require
    explicit robustness mechanisms. CI widths remain below 0.05 for all but the smallest classes, indicating that
    the observed model orderings are statistically robust.}
    \label{fig:class_iou_all}
\end{figure*}

The results in Fig.~\ref{fig:class_iou_all} establish three findings. \textbf{First}, within each training regime, Dual QAct provides broad and consistent gains over its ReLU counterpart across all three severities at the mIoU level. At Severity 2, ZS-QAct achieves an mIoU of 0.350 vs.\ 0.256 for ZS-ReLU (+36.6\% relative); Distill-QAct reaches 0.338 vs.\ 0.321 for Distill-ReLU (+5.3\%). At Severity 3, ZS-QAct improves from 0.126 to 0.194 (+54.0\%) and Distill-QAct from 0.288 to 0.310 (+7.6\%). At Severity 4, the within-regime QAct gain grows further: +66.1\% in the zero-shot regime (0.115 to 0.191) and +19.6\% in the distillation regime (0.204 to 0.244). Notably, the QAct gain within the zero-shot setting \textit{increases with severity}, reflecting that deeper blur drives more signal into the ReLU dead-zone and thereby amplifies the architectural advantage. \textbf{Second}, the gains are disproportionately concentrated in rare structural and texture-dependent classes most susceptible to ReLU statistical erasure. At Severity 4, ZS-QAct improves \textit{weed\_cluster} IoU by $4.8\times$ over ZS-ReLU (0.137 vs.\ 0.029) and \textit{water} by $2.5\times$ (0.487 vs.\ 0.197). Macro-class performance (\textit{drydown}, \textit{nutrient\_deficiency}) is preserved or improved. Some dominant classes (\textit{water}, \textit{planter\_skip}) show mixed per-class ordering between Distill-QAct and Distill-ReLU at moderate severities (2--3), consistent with blur-supervised data partially compensating for magnitude-based limitations on high-contrast classes; however, Distill-QAct remains the best overall model at all severities. This mixed class-level behavior confirms that QAct's role is to provide robust architectural inductive bias that is complementary to good blur-domain training data — not to uniformly dominate every class irrespective of training regime. \textbf{Third}, across regimes, Distill-QAct is the best overall model at all severities, and zero-shot QAct outperforms Distill-ReLU at Severity 2 as well as on \textit{water} and \textit{weed\_cluster} at Severity 4, confirming that the rank-aware inductive bias can substitute for blur-domain training data on the classes most vulnerable to statistical erasure.

\subsection{Directional Robustness: QAct Gains Are Not Direction-Specific}
\label{subsec:direction}


A potential confound in any blur-robustness claim is that performance gains may reflect a coincidental alignment between the dominant blur direction and image structure (e.g., crop rows parallel to the blur kernel reducing effective smearing). A model with genuine architectural robustness should exhibit low IoU variance across blur directions; a model benefiting from directional coincidence would show high variance. To rule this out, we evaluate all four models under directional blur at 8 angles (0°, 45°, 90°, 135°, 180°, 225°, 270°, 315°) at fixed Severity 3.

We note that the absolute mIoU values here are higher than those in the headline experiments (Section~\ref{subsec:dataset}), and the cross-regime ordering differs: ZS-QAct leads overall in the direction sweep, whereas Distill-QAct leads in the headline results. This reflects a fundamental difference in blur complexity. The direction sweep uses deterministic single-axis linear blur, a simpler degradation that QAct's rank-aware normalization handles architecturally without requiring blur-domain training. The headline experiments use randomized 6-DOF blur, whose geometric diversity across trajectories requires the learned adaptation that distillation provides. The direction sweep is therefore designed exclusively to test directional invariance — not to reproduce headline severity levels.

\begin{figure}[h]
    \centering
    \includegraphics[width=\linewidth]{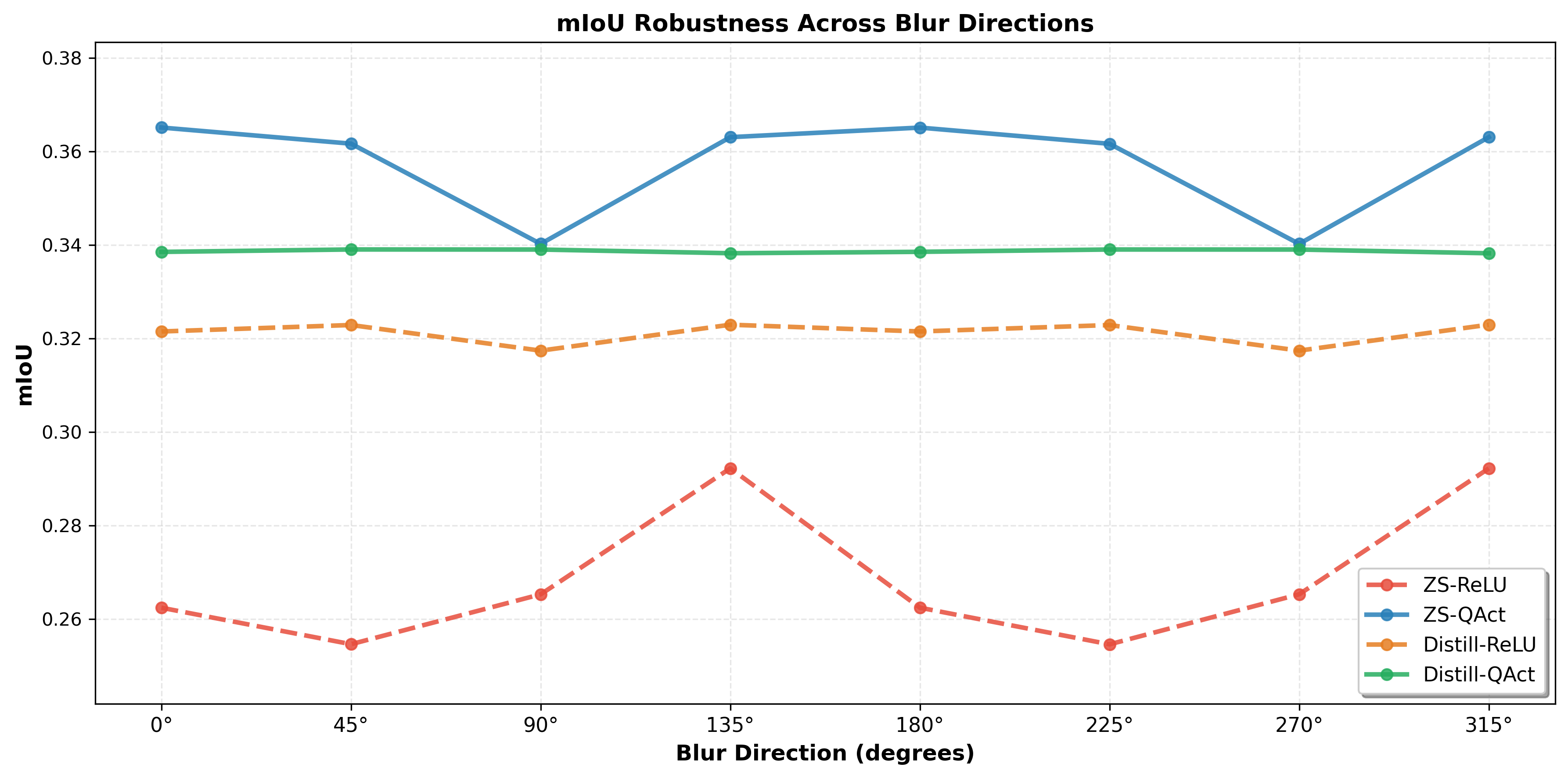}
    \caption{\textbf{Directional Robustness at Severity 3 (8 blur angles, 0°--315°).}
    Radar chart of mIoU across 8 blur directions for all four models.
    Within each training regime, QAct variants achieve both higher mean mIoU and lower
    directional variance than their ReLU counterparts: ZS-QAct (mean 0.358, std 0.010)
    vs.\ ZS-ReLU (mean 0.269, std 0.014); Distill-QAct (mean 0.339, std 0.0003)
    vs.\ Distill-ReLU (mean 0.321, std 0.002). The near-perfectly circular Distill-QAct
    profile confirms that QAct's robustness is an architectural property invariant to
    blur orientation, not an artifact of a coincidentally favored direction.}
    \label{fig:direction_sweep}
\end{figure}

\section{Conclusion}
In this letter, we addressed the gap between pristine satellite imagery and realistic, motion-blurred drone acquisitions. Motion blur physically attenuates high-frequency signals into the ReLU dead-zone, causing statistical erasure of the minority agronomic classes with the highest deployment value. By introducing Dual Quantile Activation (QAct), a rank-aware residual activation that replaces magnitude gating with instance-level rank normalization, we demonstrate that QAct is the dominant architectural factor for blur robustness: it provides consistent and growing mIoU gains over ReLU across both zero-shot and blur-supervised training and all blur severities, with particularly large margins on the rare structural and texture-dependent classes most susceptible to statistical erasure. At moderate blur, zero-shot QAct alone outperforms blur-supervised ReLU baselines, offering a practical solution for deployments where collecting blur-domain training data is infeasible. When combined with blur-supervised distillation, QAct achieves the best overall performance, confirming that rank-aware activation and domain-specific training provide complementary sources of robustness. This positions QAct as a drop-in architectural upgrade for any UAV-deployed agricultural segmentation pipeline, especially where blur training data is scarce.

\bibliographystyle{IEEEtran}
\bibliography{references}

@ARTICLE{sun2019deep,
  author={Wang, Jingdong and Sun, Ke and Cheng, Tianheng and Jiang, Borui and Deng, Chaorui and Zhao, Yang and Liu, Dong and Mu, Yadong and Tan, Mingkui and Wang, Xinggang and Liu, Wenyu and Xiao, Bin},
  journal={IEEE Transactions on Pattern Analysis and Machine Intelligence}, 
  title={Deep High-Resolution Representation Learning for Visual Recognition}, 
  year={2021},
  volume={43},
  number={10},
  pages={3349-3364},
  doi={10.1109/TPAMI.2020.2983686}}

@inproceedings{chen2018encoder,
  title     = {Encoder-Decoder with Atrous Separable Convolution for Semantic Image Segmentation},
  author    = {Chen, Liang-Chieh and Zhu, Yukun and Papandreou, George and Schroff, Florian and Adam, Hartwig},
  booktitle = {ECCV},
  year      = {2018}
}

@inproceedings{chiu2020agriculture,
  title     = {{Agriculture-Vision}: A Large Aerial Image Database for Agricultural Pattern Analysis},
  author    = {Chiu, Mang Tik and others},
  booktitle = {CVPR},
  pages     = {2828--2838},
  year      = {2020}
}

@inproceedings{chiu2020challenge,
  title     = {The 1st {Agriculture-Vision} Challenge: Methods and Results},
  author    = {Chiu, Mang Tik and others},
  booktitle = {CVPRW},
  pages     = {34--44},
  year      = {2020}
}

@article{challa2025quantile,
  title   = {Quantile Activation: Correcting a failure mode of traditional {ML} models},
  author  = {Aditya Challa and Sravan Danda and Laurent Najman and Snehanshu Saha},
  journal = {Transactions on Machine Learning Research},
  issn    = {2835-8856},
  year    = {2025},
  url     = {https://openreview.net/forum?id=nWk5OtZ7ze},
  note    = {}
}

@inproceedings{tarvainen2017mean,
  title     = {Mean Teachers are Better Role Models: Weight-Averaged Consistency Targets
               Improve Semi-Supervised Deep Learning Results},
  author    = {Tarvainen, Antti and Valpola, Harri},
  booktitle = {NeurIPS},
  volume    = {30},
  pages     = {1195--1204},
  year      = {2017}
}

@inproceedings{hinton2015distilling,
  title     = {Distilling the Knowledge in a Neural Network},
  author    = {Hinton, Geoffrey E and Vinyals, Oriol and Dean, Jeff},
  booktitle = {NeurIPS Deep Learning and Representation Learning Workshop},
  year      = {2015}
}

@inproceedings{kupyn2018deblurgan,
  title     = {{DeblurGAN}: Blind Motion Deblurring Using Conditional Adversarial Networks},
  author    = {Kupyn, Orest and Budzan, Volodymyr and Mykhailych, Mykola and Mishkin, Dmytro and Matas, Jiri},
  booktitle = {CVPR},
  pages     = {8606--8614},
  year      = {2018}
}

@inproceedings{kupyn2019deblurganv2,
  title     = {{DeblurGAN-v2}: Deblurring (Orders-of-Magnitude) Faster and Better},
  author    = {Kupyn, Orest and Martyniuk, Tetiana and Wu, Junru and Wang, Zhangyang},
  booktitle = {ICCV},
  pages     = {8878--8887},
  year      = {2019}
}

@article{xiang2024deep,
  title     = {Deep Learning in Motion Deblurring: Current Status, Benchmarks and Future Prospects},
  author    = {Xiang, Yawen and Zhou, Heng and Li, Chengyang and Sun, Fangwei and
               Li, Zhongbo and Xie, Yongqiang},
  journal   = {The Visual Computer},
  volume    = {40},
  pages     = {3909--3937},
  year      = {2024},
  publisher = {Springer}
}

@article{ulyanov2016instance,
  title   = {Instance Normalization: The Missing Ingredient for Fast Stylization},
  author  = {Ulyanov, Dmitry and Vedaldi, Andrea and Lempitsky, Victor},
  journal = {arXiv preprint arXiv:1607.08022},
  year    = {2016}
}

\end{document}